\RequirePackage{tikz}

\documentclass[pdflatex, sn-mathphys, iicol]{sn-jnl}% Math and Physical Sciences Reference Style

\usepackage[nolist]{acronym}
\usepackage{subcaption}
\usepackage{bm}
\renewcommand{\vec}[1]{\bm{#1}}
\usepackage{tikz}
\usetikzlibrary{automata, arrows.meta, positioning, decorations, calligraphy, decorations.pathreplacing, calc, backgrounds}

\definecolor{col1}{HTML}{0072BD}
\definecolor{col2}{HTML}{D95319}
\definecolor{col3}{HTML}{EDB120}
\definecolor{col4}{HTML}{7E2F8E}
\definecolor{col5}{HTML}{77AC30}
\definecolor{col6}{HTML}{4DBEEE}
\definecolor{col7}{HTML}{A2142F}
\definecolor{edgered}{HTML}{d42020}

\tikzstyle{edge} = [draw, thick, ->, shorten >=2pt]
\tikzstyle{brace style} = [ decorate, very thick, decoration = {calligraphic brace, raise=5pt, amplitude=5pt}]
\newcommand{\convexpath}[2]{
  [   
  create hullcoords/.code={
    \global\edef\namelist{#1}
    \foreach [count=\counter] \nodename in \namelist {
      \global\edef\numberofnodes{\counter}
      \coordinate (hullcoord\counter) at (\nodename);
    }
    \coordinate (hullcoord0) at (hullcoord\numberofnodes);
    \pgfmathtruncatemacro\lastnumber{\numberofnodes+1}
    \coordinate (hullcoord\lastnumber) at (hullcoord1);
  },
  create hullcoords
  ]
  ($(hullcoord1)!#2!-90:(hullcoord0)$)
  \foreach [
  evaluate=\currentnode as \previousnode using \currentnode-1,
  evaluate=\currentnode as \nextnode using \currentnode+1
  ] \currentnode in {1,...,\numberofnodes} {
    let \p1 = ($(hullcoord\currentnode) - (hullcoord\previousnode)$),
    \n1 = {atan2(\y1,\x1) + 90},
    \p2 = ($(hullcoord\nextnode) - (hullcoord\currentnode)$),
    \n2 = {atan2(\y2,\x2) + 90},
    \n{delta} = {Mod(\n2-\n1,360) - 360}
    in 
    {arc [start angle=\n1, delta angle=\n{delta}, radius=#2]}
    -- ($(hullcoord\nextnode)!#2!-90:(hullcoord\currentnode)$) 
  }
}
\jyear{2023}%

\begin{document}

\begin{acronym}
\acro{cnn}[CNN]{Convolutional Neural Network}
\acro{lraspp}[Lite R-ASPP]{Lite Reduced Atrous Spatial Pyramid Pooling}
\acro{resnet}[ResNet]{Residuale Network}
\acro{relu}[ReLU]{Rectified Linear Unit}
\acro{cam}[CAM]{Class Activation Map}

\acro{hmm}[HMM]{Hidden Markov Model}
\acro{nll}[NLL]{Negative Log Likelihood}
\acro{ece}[ECE]{Estimated Calibration Error}
\acro{map}[MAP]{Maximum A Posteriori}

\acro{vb}[VB]{Video Bronchoscopy}
\acro{icu}[ICU]{Intensive Care Unit}
\acro{emt}[EMT]{Electromagnetic Tracking}
%\acro{ent}[ENT]{Electromagnetic Navigation Bronchoscopy}
\acro{slam}[SLAM]{Simultaneous Localization And Mapping}
\acro{sfm}[SfM]{Structure from Motion}
\end{acronym}

\title[Article Title]{Airway Label Prediction in Video Bronchoscopy: Capturing Temporal Dependencies Utilizing Anatomical Knowledge}
%\title[Article Title]{Sequential Anatomically Regularized Inference of Airway Labels in Video Bronchoscopy}

\author*[1]{\fnm{Ron} \sur{Keuth}}\email{ron.keuth@student.uni-luebeck.de}

\author[1]{\fnm{Mattias} \sur{Heinrich}}%\email{mattias.heinrich@uni-luebeck.de}

\author[2]{\fnm{Martin} \sur{Eichenlaub}}%\email{martin.eichenlaub@uniklinik-freiburg.de}

\author*[1]{\fnm{Marian} \sur{Himstedt}}\email{marian.himstedt@uni-luebeck.de}

\affil*[1]{\orgdiv{Medical Informatics}, \orgname{University of L\"ubeck}, \orgaddress{\street{Ratzeburger Allee 160}, \postcode{23562} \city{L\"ubeck}, \country{Germany}}}

\affil[2]{\orgdiv{Clinic of Pneumology}, \orgname{University Hospital Freiburg, Germany}, \orgaddress{\street{Breisacher Straße 153}, \postcode{79110} \city{Freiburg}, \country{Germany}}}

%\abstract{TBA}

%%================================%%
%% Sample for structured abstract %%
%%================================%%

\abstract{\textbf{Purpose:} Navigation guidance is a key requirement for a multitude of lung interventions using video bronchoscopy. State-of-the-art solutions focus on lung biopsies using electromagnetic tracking and intraoperative image registration w.r.t. preoperative CT scans for guidance. The requirement of patient-specific CT scans hampers the utilisation of navigation guidance for other applications such as intensive care units.     

\textbf{Methods:} This paper addresses navigation guidance solely incorporating bronchosopy video data. In contrast to state-of-the-art approaches we entirely omit the use of electromagnetic tracking and patient-specific CT scans. Guidance is enabled by means of topological bronchoscope localization w.r.t. an interpatient airway model. Particularly, we take maximally advantage of anatomical constraints of airway trees being sequentially traversed. This is realized by incorporating sequences of CNN-based airway likelihoods into a Hidden Markov Model.

\textbf{Results:}  Our approach is evaluated based on multiple experiments inside a lung phantom model. With the consideration of temporal context and use of anatomical
knowledge for regularization, we are able to improve the accuracy up to to 0.98 compared to 0.81 (weighted F1: 0.98 compared to 0.81) for a classification based on individual frames.
 
\textbf{Conclusion:} We combine CNN-based single image classification of airway segments with anatomical constraints and temporal HMM-based inference for the first time. Our approach renders vision-only guidance for bronchoscopy interventions in the absence of electromagnetic tracking and patient-specific CT scans possible.  
}




\maketitle

\begin{figure*}
    \centering
    \includegraphics[width=0.9\textwidth]{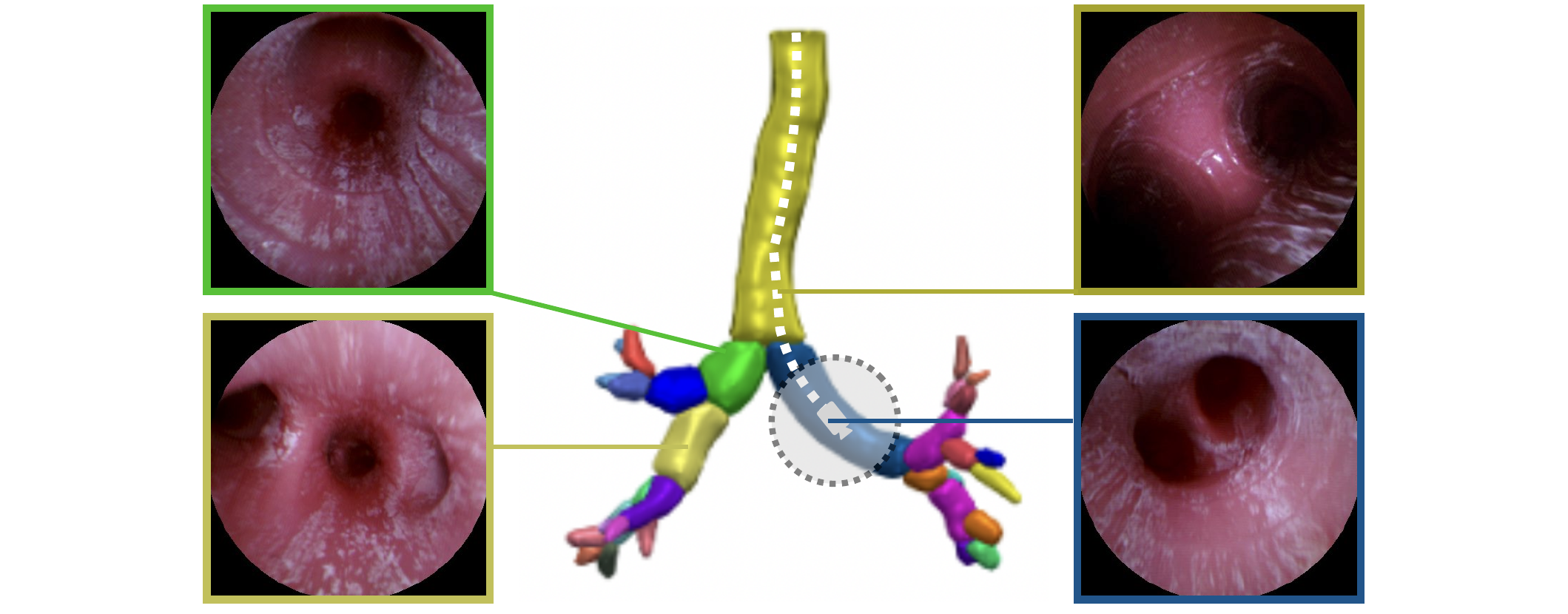}
    \caption{Interpatient model with multi-label segmentation of airways generated based on \cite{falta2022}. Brochoscopy video frames are assigned labels according to this anatomical model. Our approach predicts the airway label of the current bronchoscope location in a topological manner (grey circle; dashed line).}
    \label{fig:cover_figure}
\end{figure*}

\section{Introduction}\label{sec:intro}
\ac{vb} is frequently carried out in \acp{icu} due to several diagnostic and therapeutic indications such as removal of foreign objects, secretion sampling and suction as well as clarification of ventilation problems. 
Biopsies conducted in cases of suspected lung cancer target at a limited number of predefined locations. In contrast, \ac{vb} in \acp{icu} often require inspecting large portions of bronchial trees, i.e. interventions are performed within a large spatial extent. Constantly keeping track of the bronchoscope’s position within the airway tree poses a significant mental challenge to physicians, potentially resulting in longer treatment times, which entail increased risk for patients. This is emphasized by the fact that the majority of physicians in \acp{icu} are not pneumologists having less practical experience in bronchoscopy resulting in a greater error rate, also in the identification of upper airways \cite{yoo_deep_2021}. Today, navigation guidance is a default tool for lung biopsies, however, these systems are uncommon in \acp{icu}. The additional hardware, i.e. \ac{emt} implies substantial costs particularly due to their single-use components as well as additional setup times and requirements for medical staff \cite{eberhardt_lungpointnew_2010}. The lack of prior CT scans eventually constitutes the foreclosing reason for omitting \ac{emt} in \acp{icu} as patients are likely to be inappropriate for radiology transfers due to their unstable conditions and potential infection risks for patients and medical staff. An easy-to-use tracking solution solely utilizing interventional images and interpatient airway knowledge exhibits a beneficial tool for this application. The lung’s airways are particularly well-suited for learning a generic model as the branch variation is limited. Clinical studies \cite{smith_2018} have demonstrated $>95\%$ of patients comprise a common airway model with only two anatomical variations that up to the fourth branching generation (thereof 16\%: one accessory sub-superior segment; 6.1\%: absent right medial-basal segment) which can be adequately accounted for in the training process. A navigation beyond this level is rather uncommon for intensive care settings. The requirements to the spatial accuracy are moderate as long as the topological consistence is preserved, i.e. the correct prediction of airways and bifurcations w.r.t. an interpatient model. Robust tracking as well as preserving and highlighting traversed airways enable an increase of a physician’s confidence while simultaneously reducing intervention times and making quality less reliant on individual skills. The overall reduced risks for patients and medical staff promise a substantial impact for \acp{icu}. Omitting additional single-use parts being attached to bronchoscopes simplifies regulatory processes and minimizes costs for health insurances and other third-party payers.

This paper presents a novel approach to purely image-based navigation guidance in \ac{vb} consisting of the following components:
\begin{enumerate}
    \item A \ac{cnn}-based single-image classification of 15 airway segments up to the fourth branching generation. 
    \item An inference model for processing sequential airway likelihood data predicted by the \ac{cnn} based on \acp{hmm}.
\end{enumerate}

Our work includes a substantial calibration of the aforementioned components, ensuring optimal results which are demonstrated in exhaustive phantom experiments.

\section{Related work}\label{sec:related_work}
\paragraph{Airway classification}
In the past decade, \acp{cnn} have proven their ability to learn the extraction of task-specific features, which also enables their use in the image-based localization of the endoscope in \ac{vb}. For example, \acp{cnn} have been directly used to predict the visibility of airways as well its position and angel to the endoscope in the current frame\cite{sganga_autonomous_2019}. This approach was supplemented with a partical filter to capture the temporal context, and is thus similar to our proposed method with a \ac{hmm}. However, our method uses a \ac{cnn} that directly predicts the current location based on the frame and is then regularized within the temporal context, while Sganga's \ac{cnn} predicts only the feature for the partical filter. Navigation support by means of airway classification up to the first branching generation using CNNs is investigated by \cite{yoo_deep_2021}. Although presenting promising results and demonstrating the \ac{cnn}'s semantic understanding (by means of \ac{cam} visualizations) the benefit for navigation tasks is rather limited.

%{ Sganga et al. combine airway classification and metric pose estimation (AirwayNet) by using CNNs and subsequent particle filtering (BifurcationNet) \cite{sganga_autonomous_2019}. Even though the BifurcationNet is capable of handling interpatient airway models to a limited extent, AirwayNet is solely trained for individual cases making prior training stages for each patient indispensable. 
%}

\paragraph{Electromagnetic Navigation Bronchoscopy}
\ac{emt}-based solutions are state-of-the-art for navigation inside the lung and have been integrated by multiple commercial products, e.g. SuperDimension\textsuperscript{TM} (Medtronic Inc., Minneapolis, MN), SPiN System\textsuperscript{\textcopyright}
Veran (Veran Medical Technology Inc., St. Louis, MO) and Monarch\textsuperscript{\textcopyright} (Auris Health Inc., Redwood City, CA) utilize preoperative patient-specific CTs, EMT and video bronchoscopies for navigation guidance. Based on airway segmentations, virtual bronchoscopy images are rendered from prior CTs which are subsequently registered to intraoperative real bronchoscopy images \cite{mori2000,mori2002,nagao2004, deligianni2004}. This process is stabilized by incorporating \ac{emt}. The methodological background for this hybrid method is exhaustively investigated in \cite{reichl_2013, eberhardt_lungpointnew_2010} and  with \cite{reichl_2011} giving particular insights into deformable registration of in-vivo and virtual bronchoscopy images. Deligianni et al. investigate statistical shape models aiming at airway motion compensation in registration \cite{deligianni2006}. The utilization of interpatient knowledge is rather uncommon but exhibits superior capabilities for both, missing patient-specific CT scans and motion compensation as exhaustively motivated in \cite{smith_2018}.

\paragraph{Endobronchial pose estimation}
Solutions to endobronchial estimation of metric camera poses omitting \ac{emt} have been investigated for more than two decades. The fundamental for this is given by airway structures being segmented in CT scans. One group (a) of existing approaches generates virtual bronchoscopy sequences (2D images) that are subsequently matched to in-vivo images regarding image similarity. The particular challenge here is the domain gap arising from substantially different textures of tissue surfaces given in-vivo and rendered images respectively. To address this problem, Sganga et al. employ domain randomization across the texture space, which can mitigate the impact of textural appearances \cite{sganga_autonomous_2019}. Approaches solely examining image similarity are subject to ambiguity for global pose estimation and thus rather limited to pure (local) tracking tasks \cite{eberhardt_lungpointnew_2010}. 

\noindent Another group (b) resolves the domain gap through an intermediate representation, where depth maps are estimated from RGB images using GANs \cite{shen_context-aware_2019, zhao2019,banach2021}. Ground truth depth maps are generated based on paired CT and in-vivo datasets, which is partly accompanied by pretraining on large-scale CT datasets incorporating virtual bronchoscopies and depth maps rendered thereof. 

\noindent A third group (c) of approaches utilizes visual \ac{slam} or \ac{sfm} for metric pose estimation. Wang et al. utilize \ac{slam} based on ORB features to establish 2D-3D correspondences of intraoperative bronchoscopy frames and prior CT scans omitting \ac{emt} \cite{Wang2020_slam}. Similarly to \cite{visentini-scarzanella_deep_2017}, this approach relies on sufficient structural content to be tracked and constant camera motion. 
\noindent For an in-depth review of endobronchial pose estimation approaches, the reader is referred to \cite{borrego-carazo_bronchopose_2022}. 

\paragraph{Summary}
The majority of existing approaches focusses on the prediction of camera poses, i.e. metric navigation. While traditional approaches utilize nonlinear 2D-3D or 2D-2D registration, it can be observed that 3D-3D registration methods incorporating learned depth maps mainly acquired from virtual bronchoscopy have drawn particular interest of latest research. Predicting endobronchial locations solely from bronchoscopy sequences and thus treating pose estimation as airway recognition, i.e. topological navigation, has been barely addressed by the research community, except for the limited solution presented in \cite{yoo_deep_2021}. Most of the state-of-the-art approaches require at least prior patient-specific CT scans for navigation, which hampers their use for applications outside tissue biopsies (e.g. \ac{icu}). 

\section{Methods}\label{sec:methods}

\subsection{Dataset} \label{sec:ds}
\begin{table*}
    \centering
    \caption{Description of our defined dataset's splits and their usage. All frames of $F$ are processed independently of their sequences during the training of the \acsp{cnn}. Please read Sec. \ref{sec:ds} for further details and our motivation.}
    \resizebox{\textwidth}{!}{
    \begin{tabular}{c|l|l|c|l}
        split & processing level & sequences ID of dataset & $\#$images & used for\\\hline
        $F$ & frames & $\{3, 4, 5, 6, 8, 15\}$               & $15\,961$ & training \acsp{cnn} for classification \& segmentation \\
        $S$ & sequences  & $\{2, 9, 10, 11, 12, 13, 14, 16\}$    & $18\,463$ & optimizing $\lambda_\mathcal{R}$ (see Sec. \ref{sec:optim_lambda}) \\
        $T$ & sequences& $\{0,7\}$                              & $5175$    & test of the detection pipeline (see Fig. \ref{fig:pipeline_scheme})         
    \end{tabular}
    }
    \label{tab:dataset_splits}
\end{table*}

\begin{figure}
    \centering
    \includegraphics[width=.48\textwidth]{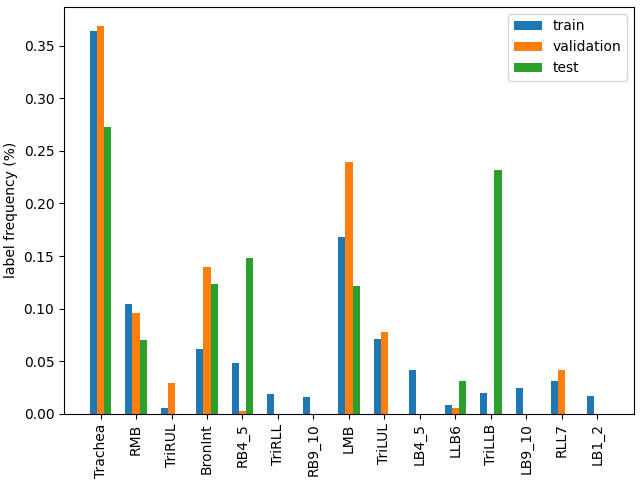}
    \caption{Class distribution of each dataset split in percent. Please see Fig. \ref{fig:baumgraph} for the anatomical position of each label.}
    \label{fig:split_class_distribution}
\end{figure}

Due to the lack of public available datasets that cover multiple sequences of in-vivo bronchoscopies, we develop our method on a synthetic but public available dataset \cite{visentini-scarzanella_deep_2017} generated on a simplified silicone phantoms of the bronchial tree.
The simplified phantom covers 17 bronchial branches (also called nodes) up to the fourth generation of the bronchial tree. Each of these nodes marks the destination for 16 bronchoscopies, with the trachea as their start- and endpoint, resulting in $39\,000$ RGB-Image in total.
The utilized phantom dataset is accompanied by a 3D mesh model of the bronchial tree, as well as the exact global position of the endoscope for each frame. However, it misses ground truth airway labels. Thus, we manually annotated the airway structures inside the mesh model enabling a subsequent automatic assignment of airways labels for individual bronchoscopy frames based on their ground truth poses w.r.t. to the mesh. \par
Because we use the dataset to optimize our \acp{cnn} as well as our \ac{hmm} subdivision into disjunct splits is crucial. We decide to split the dataset along the sequences, making sure that the split $F$ for the training of the \acp{cnn} contains all anatomical classes but as few sequences as possible, holding as many of those back for the optimization of the \ac{hmm}. We implement a greedy algorithm that starts picking the sequence holding the most uncovered anatomical classes until each class is covered, resulting in a split with six sequences and $15\,961$ images. For the final evaluation of the complete detection pipeline on unseen images and sequences, we choose the two longest sequences ($5\,175$ images), which cover the most nodes of the left and right side of the bronchial tree, as the test split $T$. Leaving eight sequences with $18\,463$ images in total for the sequence-based optimization of the \ac{hmm}. The class distribution of each final split is shown in Fig. \ref{fig:split_class_distribution} and details are summarized in Tab. \ref{tab:dataset_splits}.

\subsection{Classification pipeline}
\begin{figure*}
    \centering
    \input{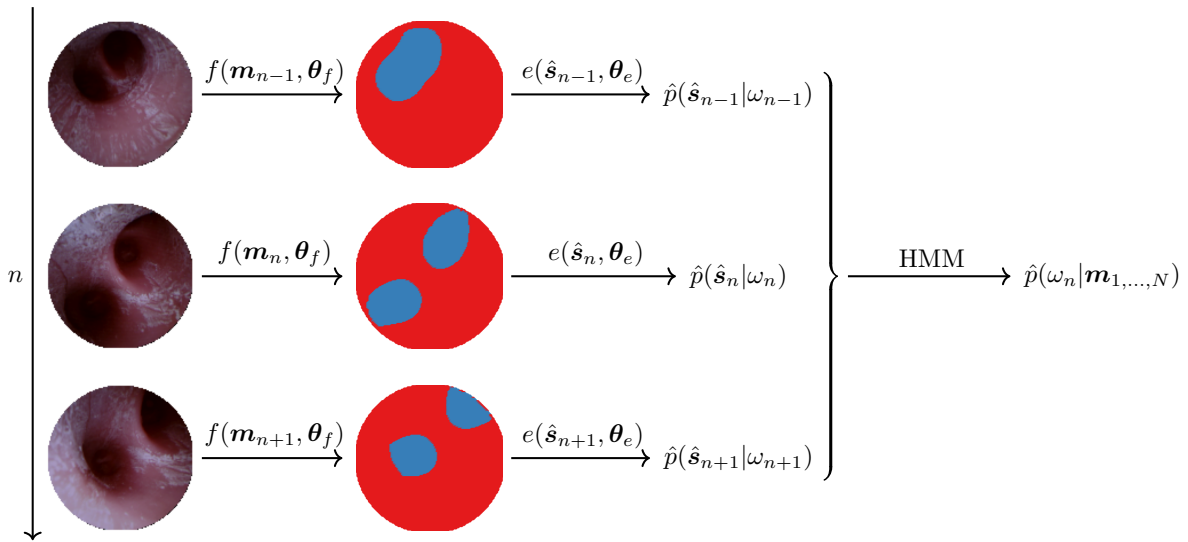}
    \caption{Structure of our proposed pipeline for the image-based localization of the endoscope during a \ac{vb}. $f$ maps the current frame $\vec{m}_{n\in[1,\dots, N]}$ to its corresponding semantic segmentation $\hat{s}_n$. The classifier $e$ predicts the likelihood $\hat{p}(\hat{s}_n\vert\omega_n)$ for each possible anatomical label $\omega$ based on $\hat{s}_n$. Finally, a \acf{hmm} captures the temporal context and predicts the posterior probability $\hat{p}(\omega_n\vert\vec{m}_{1,\dots,N})$ for the current frame given the whole sequence of frames. $f$ and $e$ are implemented via two \acp{cnn} with their trainable parameters $\vec{\theta}$.}
    \label{fig:pipeline_scheme}
\end{figure*}
Fig. \ref{fig:pipeline_scheme} shows our proposed pipeline for the image-based localization of the endoscope during a \ac{vb}. A \ac{lraspp}\cite{howard_searching_2019} generates the semantic segmentation of the bronchial orifice for each frame as an abstract scene representation to overcome image artifacts like secrete or bubbles generated by a coughing patient during the \ac{vb}.
We trained the \ac{lraspp} with a generated segmentation ground truth due to the lack of readily useable annotations.
Although this generated ground truth is just a weak supervision, we were able to show that the \ac{lraspp} gained a semantic understanding of bronchial orifices and is also fairly domain robustness \cite{keuth_weakly_2022}.
We then use the latent space of the \ac{lraspp} as the input for a shallow \ac{resnet}\cite{He2016} ($800 k$ parameters, 30 epochs, Adam with $1e-3$ learning rate, same data augmentation like in \cite{keuth_weakly_2022}) to localize the endoscope within the bronchial tree by classifying the visible anatomical structure. Due to the high class imbalance, we undersample all classes to match the less frequently one as close as possible by increasing the stride on the frames for each class individually and, in addition to that, weight the cross entropy loss for each class by its rooted inverse frequency.
Finally, a \ac{hmm} to predict the anatomical region of the current frame based on the likelihood prediction of all frames by the \ac{cnn} classifier.
The \ac{hmm} adds the missing awareness of the temporal context and enables the explicit use of anatomical a priori knowledge with its formulation within the regularization term.

\subsection{Calibration of \acs{cnn} classifier}\label{sec:calibration}
Nowadays, \acp{cnn} tend to overestimate their prediction's confidence, resulting in a gap between the (softmax-) probability associated with the predicted class and its actual likelihood, which is mainly due to the increased model capacity \cite{guo_calibration_2017}. However, as the likelihood prediction defines the base of our \ac{hmm}, this miscalibration would bias the classification within the temporal context and is therefore addressed by the temperature scaling \cite{guo_calibration_2017}, which introduces a variable $T\in\mathbb{R}^+$ to globally scale the logits $\vec{z}$ of the model.
\begin{equation}\label{eq:calibrated_l_prediction}
    \hat{p}(\vec{m}\vert\omega_i) = \sigma_{\text{SM}}^{(i)}\left(\frac{\vec{z}}{T}\right)
\end{equation}
where $\sigma_{\text{SM}}^{(i)}$ describes the softmax score for the $i$-th class.
$T$ is optimized using the \ac{nll} to fit the model's prediction distribution to the ground truth distribution of the validation data with initial $T_0=1$.

% The validation set is used to calculate a function which maps the confidence of the model to its actual accuracy. If a model is perfectly calibrated, this function is the identity function. Due to the limited number of samples in the validation set, the value range of the confidence $[0, 1]$ is devided into $m$ bins. This enables the derivation from the identity function with the \acf{ece}\cite{guo_calibration_2017}:
% \begin{align}
%     \acs{ece}&=\sum_m^M\frac{\vert B_m\vert}{n}\vert \text{acc}(B_m) - \text{conf}(B_m)\vert\\
%     \text{acc}(B_m)&=\frac{1}{\vert B_m\vert}\sum_{i\in B_m}1(\hat{y}_i=y_i)\\
%     \text{conf}(B_m)&=\frac{1}{\vert B_m\vert}\sum_{i\in B_m}\hat{p}_i
% \end{align}
% where $B_m$ is the set containg all sample of the $m$-th bin. conf the confidence of the model and acc the accuracy with the label $y_i$ for the $i$-th class.

\subsection{Dynamic programming for time domain}
\subsubsection{\acl{hmm}}
A \ac{vb} can be considered as a sequence of frames $\{\vec{m}_n\}_{n=1}^N$ with their likelihood probability distribution $p(\vec{m}\vert\omega)$ for each anatomical class $\omega\in\Omega$, where the likelihood is predicted by the calibrated \ac{cnn} classifier. The \acf{hmm} models the temporal context within this sequence, obtaining the most likely label sequence via a \ac{map} estimation.
    \begin{equation}
        \resizebox{0.37\textwidth}{!}{$\hat{\omega}_{1,\dots,N}=\underset{\omega_{1,\dots,N}}{\arg\max}\ p(\vec{m}_{1,\dots,N}\vert\omega_{1,\dots,N})p(\omega_{1,\dots,N})$}
    \end{equation}
    \begin{equation}
    \resizebox{0.47\textwidth}{!}{$\hat{\omega}_{1,\dots,N}=\underset{\omega_{1,\dots,N}}{\arg\max}\left[\left(\prod_{n=1}^Np(\vec{m}_n\vert\omega_n)\right)\left(\prod_{n=2}^Np(\omega_n\vert\omega_{n-1})\right)\right]\label{eq:hmm}$}
    \end{equation}

% original equation (align env) 
%    \begin{align*}
%    \hat{\omega}_{1,\dots,N}&=\underset{\omega_{1,\dots,N}}{\arg\max}\ p(\vec{m}_{1,\dots,N}\vert\omega_{1,\dots,N})p(\omega_{1,\dots,N})\\
%    &=\underset{\omega_{1,\dots,N}}{\arg\max}\left[\left(\prod_{n=1}^Np(\vec{m}_n\vert\omega_n)\right)\left(\prod_{n=2}^Np(\omega_n\vert\omega_{n-1})\right)\right]\label{eq:hmm}
%    \end{align*}

where Eq. \ref{eq:hmm} holds due to the \ac{hmm} modelling the prior of each timestamp as the transition probability between different classes in $\Omega$. A more generalized formulation considers the likelihood as a unitary data term $\mathcal{D}_n(\omega_n)$ as well as the prior as a pairwise regularization term $\mathcal{R}_n(\omega_n,\omega_{n-1})$
\begin{equation}\label{eq:hmm_generalized}
    \resizebox{0.47\textwidth}{!}{$\hat{\omega}_{1,\dots,N}=\underset{\omega_{1,\dots,N}}{\arg\min}\left[\sum_{n=1}^N\mathcal{D}_n(\omega_n)+\sum_{n=2}^N\mathcal{R}_n(\omega_n,\omega_{n-1})\right]$}
\end{equation}
with all probabilities considered as negated and logrithmized \cite[Sec. 11.1]{prince_computer_2012}.\par

\subsubsection{Data and regularization term}
\begin{figure}
    \centering
    \resizebox{!}{.5\textwidth}{
        \begin{tikzpicture}[auto, scale=1]
    \node (trachea) [state, initial, initial above] {Trachea};
    
    \node (lmb) [state, below right = of trachea] {LMB};
    \node (trilul) [state, right = of lmb] {TriLUL};
    \node (lb45) [state, below right = of trilul] {LB4+5};
    \node (lb12) [state, above right = of trilul] {LB1+2};
    \node (llb6) [state, below = of lmb] {LLB6};
    \node (trillb) [state, below = of llb6] {TriLLB};
    \node (lb910) [state, below = of trillb] {LB9+10};

    \node (rmb) [state, below left = of trachea] {RMB};
    \node (trirul) [state, left = of rmb] {TriRUL};
    \node (bronint) [state, below left = of rmb] {BronInt};
    \node (rb45) [state, below right = of bronint] {RB4+5};
    \node (rll7) [state, below = of bronint] {RLL7};
    \node (trirll) [state, below = of rll7] {TriRLL};
    \node (rb910) [state, below = of trirll] {RB9+10};
    
    \path [thick]
        (trachea) edge (lmb)
        (lmb) edge (trilul)
        (trilul) edge (lb12)
        (trilul) edge (lb45)
        (lmb) edge (llb6)
        (llb6) edge (trillb)
        (trillb) edge (lb910)

        (trachea) edge (rmb)
        (rmb) edge (trirul)
        (rmb) edge (bronint)
        (bronint) edge (rb45)
        (bronint) edge (rll7)
        (rll7) edge (trirll)
        (trirll) edge (rb910)
        ;
\end{tikzpicture}
    }
    \caption{An undirectional tree graph modelling the bronchial tree with labeled bronchial branches (nodes) covered by our phantom (see Fig. \ref{fig:cover_figure}). For simplicity, the distance between adjacent bronchial branches is 1 regardless of their actual anatomical distance.}
    \label{fig:baumgraph}
\end{figure}
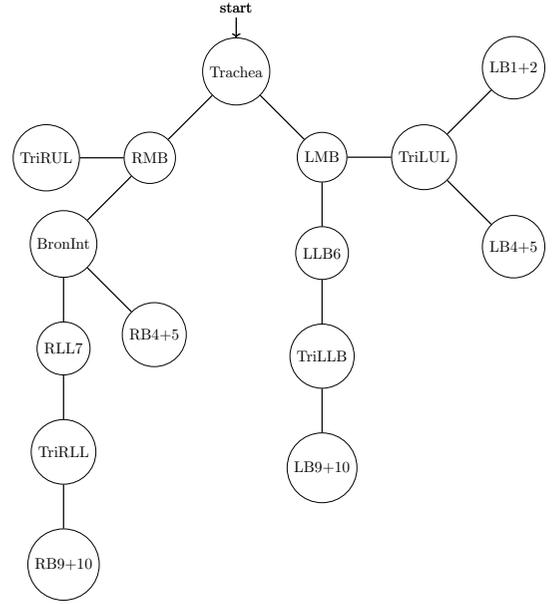
The data term models the likelihood $p(\vec{m}_n\vert\omega_n)$ of an anatomical region $\omega$ is visible in the current frame $\vec{m}_n$. Our \ac{cnn} classifier can be directly optimized with the \ac{nll} to predict this likelihood. After its calibration, the prediction likelihood $\hat{p}(\vec{m}\vert\omega_i)$ of Eq. \ref{eq:calibrated_l_prediction} is finally reformulated to match the minimization style of Eq. \ref{eq:hmm_generalized}:
\begin{equation}
    \mathcal{D}_n(\omega_i)=\frac{1-\hat{p}(\vec{m}\vert\omega_i)}{\vert\Omega\vert-1}.
\end{equation}
We initialize $\mathcal{D}_0$ and $\mathcal{D}_N$ with the one-hot vector for the class trachea because every sequence in our database will start and end there.\par

The generalized formulation of the regularization term as an arbitrary cost term and enables the explicit use of anatomical a priori knowledge. In the context of \ac{vb}, the cost is represented as the distance between the bronchial branches to explicitly model the anatomical knowledge. Therefore, we model the bronchial tree as a tree graph with the trachea as its root, but consider it as an undirected graph apart from its definition \cite[Sec. 2.4]{aho_design_1974} (see Fig. \ref{fig:baumgraph}). We precompute the distance matrix $\vec{D}\in\mathbb{N}_0^{\vert\Omega\vert\times\vert\Omega\vert}$ with a simple depth-first search and normalize it by its maximum.
\begin{equation}\label{eq:regularisierung}
    \mathcal{R}(\omega_i,\omega_j)=\exp\left(\frac{\vec{D}_{ij}}{\max(\vec{D}_{ij})}\right)
\end{equation}
This kind of formulation penalizes any rapid label changes to a distant region that are not plausible within the anatomy of the bronchial tree, depending on its distance.

\subsubsection{Viterbi algorithm}
Calculating every possible sequence to find the most likely one has an exponential runtime complexity $\mathcal{O}(\vert\Omega\vert^N)$ and is therefore not feasible for longer sequences like the number of frames of a \ac{vb} in our use case. The Viterbi algorithm reduces the runtime complexity to a polynomial one $\mathcal{O}(N\vert\Omega\vert^2)$ implementing a dynamic programming approach. This approach breaks up this global minimization problem into multiple local minimization problems. This is achieved by passing the local solution of one sequence step to the following one via message passing. In this way, the solution of the next step is calculated recursively on the solution of the previous step:
\begin{equation}
\resizebox{0.47\textwidth}{!}{
    $m_n(\omega_n)=\mathcal{D}_n(\omega_n) + \underset{\omega_{n-1}}{\min}[m_{n-1}(\omega_{n-1})+\lambda_\mathcal{R}\mathcal{R}(\omega_n,\omega_{n-1})]
    $}
\end{equation}
with $m_0(\omega_n)=\mathcal{D}_0(\omega_n)$ and where $\lambda_\mathcal{R}\in\mathbb{R}_0^+$ describes the weighting of the data and regularization term as a hyperparameter \cite[Sec. 11.2.1]{prince_computer_2012}.

\subsubsection{Approximation of the forward-backward algorithm}
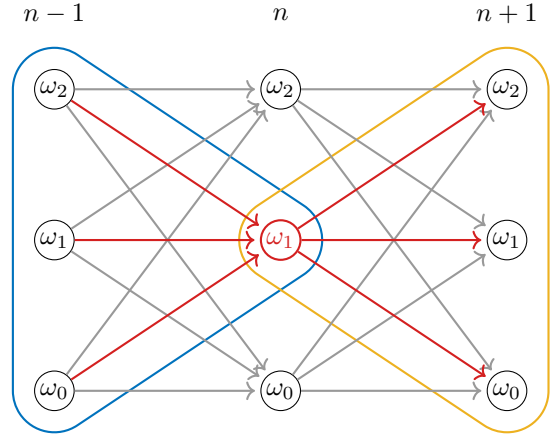
\begin{figure}
    \tikzstyle{red edge} = [edge, draw=edgered]
    \tikzstyle{gray edge} = [draw=black!40, edge]
    \tikzstyle{vertex} = [circle, minimum size=12pt, inner sep=1pt, draw]
    \tikzstyle{red vertex} = [vertex, draw=edgered, thick]
    \tikzstyle{hull} = [thick]
     
    \centering
    \begin{tikzpicture}[scale=1.0]
    % Draw vertices
    \foreach \name/\pos/\lab in {
        {s11/(0,0)/$\omega_2$},{s12/(3,0)/$\omega_2$},{s13/(6,0)/$\omega_2$},
        {s21/(0,-2)/$\omega_1$},{s23/(6,-2)/$\omega_1$},
        {s31/(0,-4)/$\omega_0$},{s32/(3,-4)/$\omega_0$},{s33/(6,-4)/$\omega_0$}}
        \node[vertex] (\name) at \pos {\lab};

    % Draw middle vertex
    \node[red vertex] (s22) at (3,-2) {\textcolor{edgered}{$\omega_1$}};

    % Draw labels
    \foreach \name/\pos/\lab in {
        {n-1/(0,1)/$n-1$},{n/(3,1)/$n$},{n+1/(6,1)/$n+1$}}
        \node (\name) at \pos {\lab};

    % Draw black arrows
    \foreach \source in {s11,s21,s31}
        \foreach \dest in {s12,s32}
            \draw[gray edge] (\source) -- (\dest);
    \foreach \source in {s12,s32}
        \foreach \dest in {s13,s23,s33}
            \draw[gray edge] (\source) -- (\dest);

    % Draw red arrows
    \foreach \source in {s11,s21,s31}
        \draw[red edge] (\source) -- (s22);
    \foreach \dest in {s13,s23,s33}
        \draw[red edge] (s22) -- (\dest);

    \begin{scope}[on background layer]
        % Draw hulls 
        \draw [draw=col1, hull] \convexpath{s22,s31,s11}{0.55cm};
        \draw [draw=col3, hull] \convexpath{s22,s13,s33}{0.55cm};
    \end{scope}
\end{tikzpicture}
    \caption{Intuition for the approximation of the forward-backward algorithm via two Viterbi, enabling the calculation of the marginal distribution over all classes being proportional to the posterior probabilities. The $n\in N$ individual steps of the sequence with their possible labels $\omega\in\Omega$ are shown from left to right.  All paths considered by the forward-backward algorithm are drawn in red. The blue hull marks the incoming paths covered by the Viterbi $m_n^f(\omega_1)$ running forward through the sequence, and the orange hull the one running backwards and covering all outgoing paths $m_n^b(\omega_1)$.}
    \label{fig:viterbi_marginal}
\end{figure}

The forward-backward algorithm\cite[Sec. 11.4]{prince_computer_2012} calculates the sum of all paths through $\omega_n$ to obtain not only the most likely class label for each time step, but also the probability distribution over all classes:
\begin{equation}
\resizebox{0.47\textwidth}{!}{    $p(\omega_n\vert\vec{m}_{1,\dots,N})\propto p(\vec{m}_{1,\dots,n}\vert\omega_n) p(\omega_n)p(\vec{m}_{n+1,\dots,N}\vert\omega_n)$}
\end{equation}
This right hand of the equation can be intuitively implemented via two Viterbi: One starting at the beginning of the sequence and calculates the most likely path up to $\omega_n$ to approximate $p(\vec{m}_{1,\dots,n}\vert\omega_n)$ with $m_n^f(\omega_n)$. The other approximates $p(\vec{m}_{n+1,\dots,N}\vert\omega_n)$ starting at the last times step of the sequence and going backwards up to $\omega_n$. Fig. \ref{fig:viterbi_marginal} visualizes the schema of this intuition.

Because both Viterbi consider the $p(\omega_n)$, it has to be subtracted ones to obtain the correct marginal distributions proportional to the posterior distribution (see Fig.\ref{fig:viterbi_marginal}). However, due to the use of the minimum and not the sum as an aggregation function, the resulting marginal distribution can only be considered as an approximation of the distribution obtained by the forward-backward algorithm.
\begin{equation}
    p(\omega_n\vert\vec{m}_{1,\dots,N})\propto m_n^f(\omega_n)+m_n^b(\omega_n)-\mathcal{D}_n(\omega_n)
\end{equation}

\subsubsection{Optimization of $\lambda_\mathcal{R}$}\label{sec:optim_lambda}
We normalize the data as well as the regularization term to a comparable value range to avoid that the weighting $\lambda_\mathcal{R}$ between those terms has to deal with too different scales and can therefore be only a semantic one. We use gradient descend to optimize $\hat{\lambda}_\mathcal{R}^\text{GD}$ over the eight sequences of the validation split. This is possible due to the provided marginal class distributions for each time step, that can be compared to the ground truth using the \ac{nll}. We init $\hat{\lambda}_\mathcal{R}^\text{GD}=1$ to start with an equal weighting for the data and regularization term. To guarantee the value space of $\mathbb{R}_0^+$, we activate $\hat{\lambda}_\mathcal{R}^\text{GD}$ with a \ac{relu}. The optimization of only one parameter enables the use of a memory-expensive second order optimizer like the L-BFGS\cite{liu_limited_1989}, resulting in a short training time with few iterations necessary. To evaluate the results, we also obtain a minimum by a brute force search within a reasonable interval considering the \ac{nll} plot (see Fig. \ref{fig:viterbi_lambda_search_interval}) with 240 samples.

\section{Results}\label{sec:results}
\begin{figure}
    \centering
    \begin{subfigure}[t]{.45\textwidth}
        \includegraphics[trim={0.7cm, 0.2cm, 1.5cm, 0.8cm}, clip, width=\textwidth]{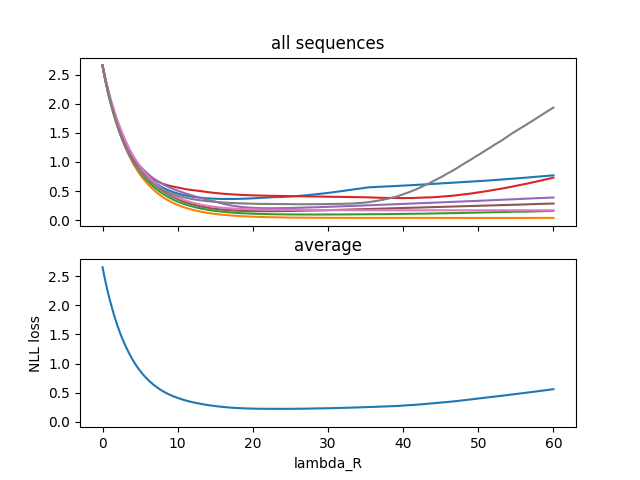}
        \caption{\acl{nll}}
    \end{subfigure}\hspace{0.04\textwidth}%
    \begin{subfigure}[t]{.45\textwidth}
        \includegraphics[trim={0.7cm, 0.2cm, 1.5cm, 0.8cm}, clip, width=\textwidth]{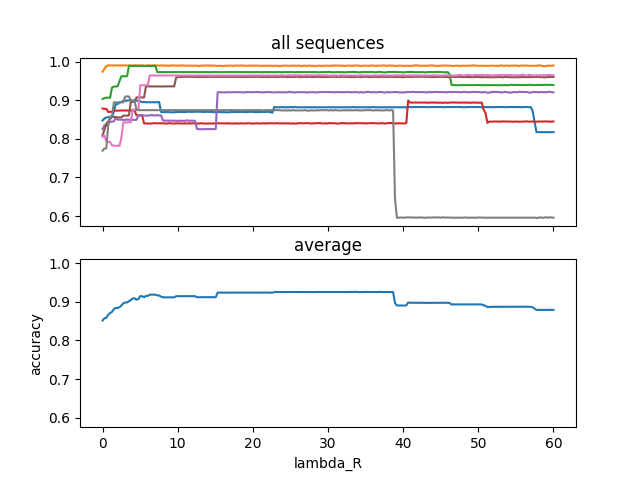}
        \caption{Accuracy}
    \end{subfigure}
    \caption{The search space covered by the brute-force method to find the optimal weighting $\hat{\lambda}_\mathcal{R}\in[0, 60]$ between data and regularization term. The upper figure shows the \ac{nll} and accuracy of the eight validation sequences, which are used for the optimization, the lower plot their average. The minimum found with gradient descent $\hat{\lambda}_\mathcal{R}^\text{GD}=22.43$ sufficiently approximates the one found via brute-force search $\hat{\lambda}_\mathcal{R}^\text{BF}=23.5$ considered the gradient of \ac{nll} in this area.}
    \label{fig:viterbi_lambda_search_interval}
\end{figure}

The Viterbi enables the optimization of the weighting of data and regularization term $\lambda_\mathcal{R}$ on the eight validation sequences of the dataset. The normalization of the data and regularization term value range to a comparable one of $[0, 1]$ is a crucial step to stabilize this optimization. However, the additional exponential amplification of the regularization term after the normalization is necessary to prevent the label change between adjacent frames. Starting with $\hat{\lambda}_\mathcal{R}^\text{GD}=1$ the optimization using the L-BFGS and \ac{nll} find a minimum at $\hat{\lambda}_\mathcal{R}^\text{GD}=22.43$. We validate this result with the minimum obtained by the brute force search with $\hat{\lambda}_\mathcal{R}^\text{BF}\in[0,60]$ and 240 samples resulting in a step size of $0.25$ and a $\hat{\lambda}_\mathcal{R}^\text{BF}=23.5$. Fig. \ref{fig:viterbi_lambda_search_interval} shows the \ac{nll} and accuracy of the eight individual sequences as well as their average for this search space.\par

We evaluate the detection pipeline with the $\hat{\lambda}_\mathcal{R}^\text{GD}$ on the two unused sequences of test split $T$. Those sequences are manually chosen and cover the lower lopes of the left and right bronchial tree. We compare the performance of the frame-based classification (only \ac{cnn} classification) with the one from the Viterbi. With the consideration of temporal context and use of anatomical knowledge for regularization, the accuracy for the first sequence can be improved by 17 points from $0.8061$ to $0.9775$ as well as for the second by 5 points from $0.6159$ to $0.6649$.\par
\begin{figure*}
    \centering
    \begin{subfigure}{\textwidth}
        \centering
        \begin{subfigure}{.45\textwidth}
            \includegraphics[trim={0.3cm, 0.8cm, 1.9cm, 0cm}, clip, width=\textwidth]{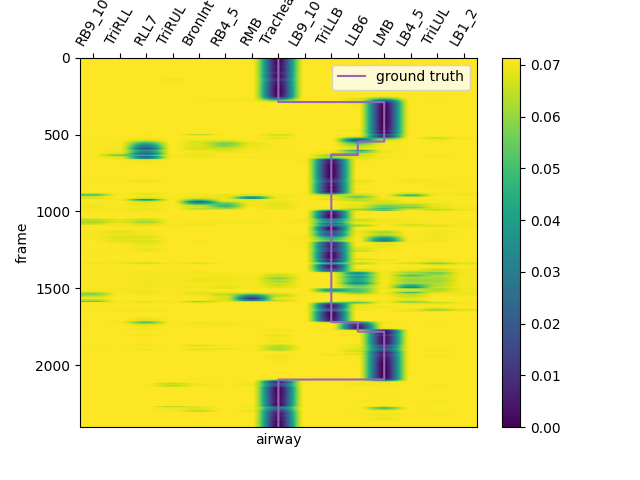}
        \end{subfigure}\hspace{0.04\textwidth}%
        \begin{subfigure}{.45\textwidth}
            \includegraphics[trim={0.3cm, 0.8cm, 1.9cm, 0cm}, clip, width=\textwidth]{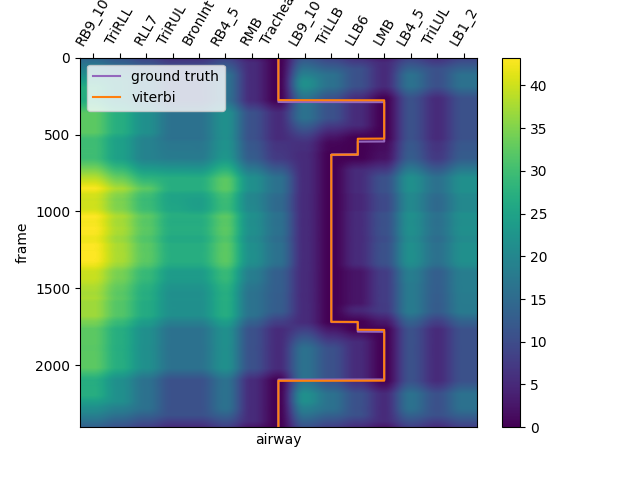}
        \end{subfigure}
        \caption{First test sequence: The Viterbi improves the accuracy of the frame-based classification from $0.8061$ to $0.9775$ and reduces the mean distance of $\vec{D}$ within the bronchial tree model from $1.09 \pm 2.02$ to $0.34 \pm 0.47$.}
        \label{fig:viterbi_ph0}
    \end{subfigure}
    \newline
    %\vspace*{.6cm}
    %\newline
    \begin{subfigure}{\textwidth}
        \centering
        \begin{subfigure}{.45\textwidth}
            \includegraphics[trim={0.3cm, 0.8cm, 1.9cm, 0cm}, clip, width=\textwidth]{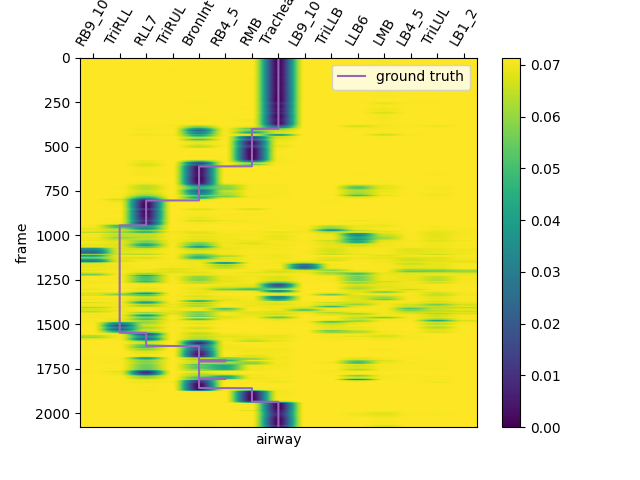}
        \end{subfigure}\hspace{0.04\textwidth}%
        \begin{subfigure}{.45\textwidth}
            \includegraphics[trim={0.3cm, 0.8cm, 1.9cm, 0cm}, clip, width=\textwidth]{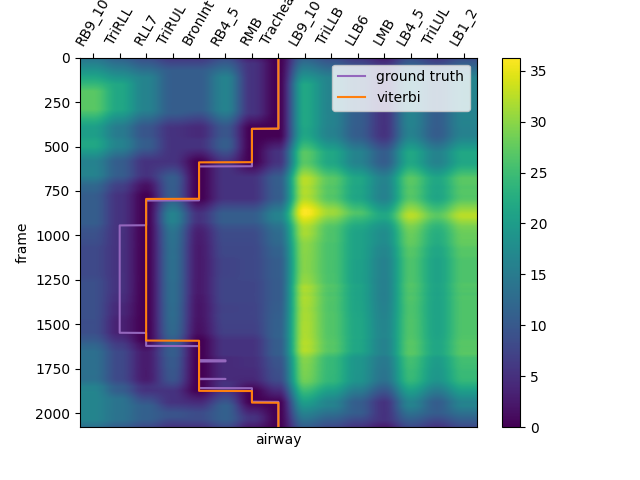}
        \end{subfigure}
        \caption{Second test sequence: The Viterbi improves the accuracy of the frame-based classification from $0.6159$ to $0.6649$ and reduces the mean distance of $\vec{D}$ within the bronchial tree model from $0.66 \pm 1.62$ to $0.02\pm 0.15$.}
        \label{fig:viterbi_ph1}
    \end{subfigure}
    \caption{Visualization of the cost for each airway label and frame. The left plots show the frame-based classification (calibrated \ac{cnn}), which is also used as the data term for the Viterbi. The right column shows the classification with the temporal awareness and anatomical regularization implemented with the Viterbi. The predicted path and the ground truth are highlighted. Due to the rapid change of labels between adjacent frames, the predicted path is omitted for the frame-based classification to improve readability. However, the minimum-cost path can be extracted by the color coding of the plot.}
    \label{fig:viterbi_cost_plot}
\end{figure*}

Tab. \ref{tab:quantitaive_ergebnisse} shows a collection of common classification metrics along with the average distance calculated based on matrix $\vec{D}$ (see Eq. \ref{eq:regularisierung}) comprising the shortest paths airways within our bronchial tree model (see Fig. \ref{fig:baumgraph}). 
For a qualitative comparison, we visualize the cost for each airway label and frame (see Fig. \ref{fig:viterbi_cost_plot}) for the frame-based \ac{cnn} classification as well as the Viterbi. We rearrange the labels according to their anatomical distance within the bronchial tree (collapsing the tree on to a line). This enables the interpretation of the cost over time as the path taken by the endoscope within the phantom of the bronchial tree. We highlight the predicted path as well as the ground truth.

\begin{table*}
    \centering
    \caption{Quantitative results on the two test sequence of our classifier without and with temporal awareness. All metrics use a micro manner weighting to compensate the high class imbalanced and to emphasize the overall performance over all classes equally. $\vec{D}$ describes the distance of the predicted and the ground truth airway label in our bronchial tree graph (please see Fig. \ref{fig:baumgraph} and Eq. \ref{eq:regularisierung} for further details). Acc@3 and especially $\vec{D}$ emphasize the superiority of the Viterbi with its time awareness and anatomically regularization, showing that the correct airway label is at least under the three most probable predictions and in the very anatomical neighborhood of the current one. }
    \begin{tabular}{cl|cccccccc}
    seq ID & classifier & Acc@1 & Acc@3 & Precision & Recall & F1 & AUC & $\vec{D}(\mu\pm\sigma)$ \\\hline
    \multirow[c]{2}{*}{0} & frame-based & 0.81 & 0.88 & 0.81 & 0.81 & 0.81 & 0.95 & $0.66 \pm 1.62$ \\
     & viterbi & 0.98 & 1.00 & 0.98 & 0.98 & 0.98 & 1.00 & $0.02 \pm 0.15$ \\ \hline
    \multirow[c]{2}{*}{1} & frame-based & 0.62 & 0.76 & 0.62 & 0.62 & 0.62 & 0.91 & $1.09 \pm 2.02$ \\
     & viterbi & 0.66 & 1.00 & 0.66 & 0.66 & 0.66 & 0.90 & $0.34 \pm 0.47$ \\ \hline \hline
    \multirow[c]{2}{*}{average} & frame-based  & 0.71 & 0.82 & 0.71 & 0.71 & 0.71 & 0.93 & $0.87 \pm 1.82$ \\
    & Viterbi & 0.82 & 1.00 & 0.82 & 0.82 & 0.82 & 0.95 & $0.18 \pm 0.31$ \\
    \end{tabular}
    \label{tab:quantitaive_ergebnisse}
\end{table*}

\section{Discussion}\label{sec:discussion}
The multiple use of the split of the data set (see Tab. \ref{tab:dataset_splits}) could course an unnoticed bias in the evaluation of our method. However, we have been limited due to the lack of other public available data sets.
The approximation of the forward-backward algorithm by applying the Viterbi forwards and backwards on the data enables the visualization of the next likely path as well as the confidence of the algorithm. It also additionally enables the direct optimization of $\hat{\lambda}_\mathcal{R}$ on the data set, which makes the manually tweaking of this hyperparameter obsolete. The result $\hat{\lambda}_\mathcal{R}^\text{GD}=22.43$ using gradient descend is comparable to the one found using a brute-force search $\hat{\lambda}_\mathcal{R}^\text{BF}=23.5$, which explicitly holds considering the gradient of the \ac{nll} in this area (see Fig. \ref{fig:viterbi_lambda_search_interval}). To evaluate the generalization power of this method, but especially the estimated $\hat{\lambda}_\mathcal{R}$, additional datasets including in-vivo \ac{vb} are needed. One other interesting aspect that has to be evaluated is the sensitivity of $\hat{\lambda}_\mathcal{R}$ to different frame rates.\par

It has to be mentioned that our airway label generation could lead to a noisy ground truth, especially at the passing to another airway, explaining the ground truth “hick ups” in the second sequence around the frame index 1700 (see Fig. \ref{fig:viterbi_ph1}).
The Viterbi overall improves the classification and all six metrics of the quantitative results shown in Tab. \ref{tab:quantitaive_ergebnisse} by introducing the temporal context. The use of anatomical knowledge as a regularization prevents implausible jumps within a few frames. For example, without the Viterbi the prediction by the \ac{cnn} would flicker from the left lower lope of the bronchial tree to the right lower lope (see Fig. \ref{fig:viterbi_ph0} at frame index 600 and 1600) and vice versa (see Fig. \ref{fig:viterbi_ph1}) covering long implausible anatomical distances like between the “LLB6” and “RLL7”. So, the prediction of the \ac{cnn} get successfully regularized to the most likely sequence that covers an anatomical plausible path within in the bronchial tree. This is also emphasized by the increase of the top 3 accuracy (Acc@3) from an average of $0.82$ to $1$, and particularly with the decrease of the average distance and its standard derivation within the bronchial tree graph from $0.87 \pm 1.82$ to $0.18 \pm 0.31$. This demonstrates the benefits of regularizing prediction w.r.t. anatomical constraints, i.e. favouring adjacent airways in the bronchial tree.\par

A disadvantage of this method is the introduced inertia in changing to a new label, which is comparable to the group delay of a common low pass filter. This results in the need for a long sequence of frames with a fairly high confidence for a specific label. An example of such a situation can be seen in Fig. \ref{fig:viterbi_ph1} around the frame index 1500: Even if the \ac{cnn} shows a high confidence for the correct label “TriRLL”, the Viterbi's prediction stays to “RLL7” due to the only short appearance of “TriRLL” and the associated additional and expensive changes in class. Even if this behavior is acceptable and a possible sequence for a \ac{vb}, a much frequently label change would be preferable, especially consider the real-world application, where it is a common use case for the physician to quickly check an airway.

\section{Conclusion}\label{sec:conclusion}
In this work, we proposed a classification pipeline for an image-based localization of the endoscope during a \ac{vb}. Our pipeline consists of multiple steps using a semantic bronchial orifice segmentation as an abstract scene representation and a \ac{cnn} that classifies the anatomical region based on the segmentation. We demonstrated that the use of a Viterbi can successfully capture the temporal context and enables the explicit use of anatomical knowledge to force the classification to sequences being plausible within the bronchial tree.
The approximation of the posterior distribution by applying the Viterbi forward and backward gives good insights on the confidence of the classification as well as the next likely alternative sequence. However, the Viterbi also introduces an inertia, requiring longer sequences with highly confident predictions to change to this class. 
Based on our outstanding results obtained from a phantom study, we will investigate the generalization capabilities of this approach to in-vivo datasets. Note, that (to our best knowledge) this is the first-time approach to vision-only bronchoscopy guidance using generic anatomy-regularized topological airway navigation omitting electromagnetic tracking and patient-specific CT scans. This is a first important step towards future bronchoscopy guidance enabling deployments also outside lung biopsies, e.g. at intensive care units.

\section*{Declarations}

Some journals require declarations to be submitted in a standardised format. Please check the Instructions for Authors of the journal to which you are submitting to see if you need to complete this section. If yes, your manuscript must contain the following sections under the heading `Declarations':

\begin{itemize}
\item Funding
\item Conflict of interest/Competing interests (check journal-specific guidelines for which heading to use)
\item Ethics approval 
\item Consent to participate
\item Consent for publication
\item Availability of data and materials
\item Code availability 
\item Authors' contributions
\end{itemize}

\noindent
If any of the sections are not relevant to your manuscript, please include the heading and write `Not applicable' for that section.

\bibliography{references}% common bib file


\end{document}